Temporal Stability and Few-Shot Prompting in Math Task Assessment

Danielle S. Fox[1], Brenda L. Robles[2], Elizabeth DiPietro Brovey[2], Christian D. Schunn[1,2]

[1]Learning Research and Development Center, University of Pittsburgh

[2]Institute for Learning, University of Pittsburgh

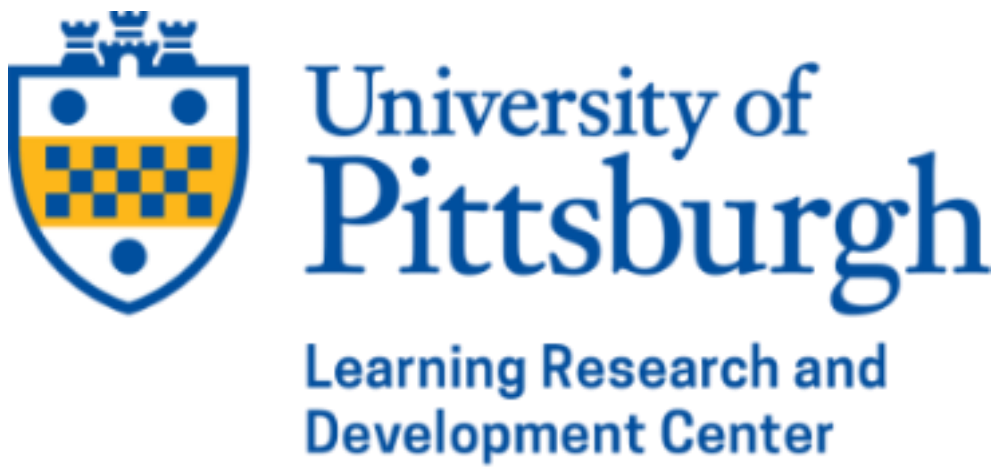

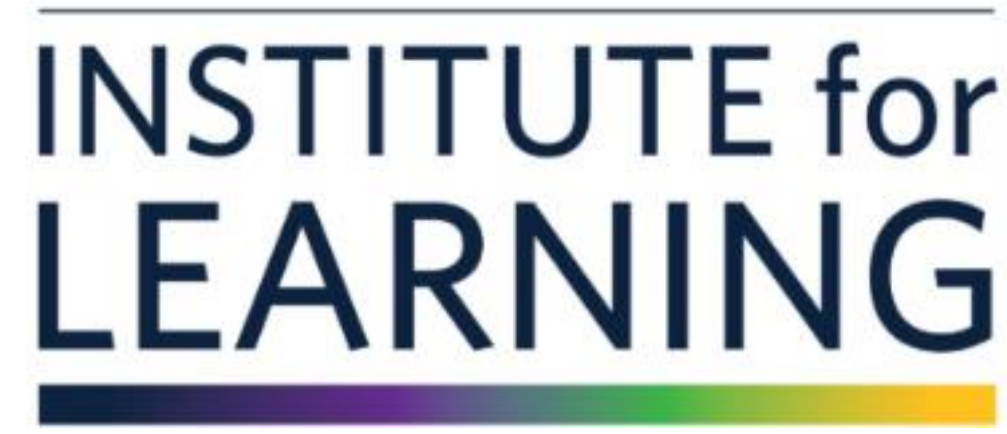

**Abstract**

As AI tools become increasingly integrated into educational contexts, questions arise about both their stability over time and their responsiveness to prompt engineering techniques. This longitudinal study focused on different AI tools' ability to use the Task Analysis Guide (TAG; Stein & Smith, 1998) to classify the cognitive demand of mathematics tasks. In particular, it examined whether this classification ability changed with (1) model version updates over time and (2) few-shot prompting using exemplar tasks. We tested a general-purpose AI tool (Gemini) and an education-specific AI tool (Coteach). The specific tools were selected because of their relatively high performance on relevant published benchmarks and prior task-specific tests. Models were tested at baseline, retested with model version updates, and then tested again using few-shot prompting (two exemplar tasks for each cognitive demand category). Results revealed that newer model versions alone produced mixed effects: Gemini's accuracy remained stable at 58%, while Coteach's accuracy decreased from 75% to 50%. However, few-shot prompting improved both models' performance: Gemini increased to 67% and Coteach recovered to 75% accuracy. These findings demonstrate that prompt engineering techniques can have larger and more reliable effects than passive model improvements, and that version updates may not always improve performance on specialized educational tasks. The study has important implications for how educators and researchers should approach AI tool selection, evaluation, and implementation in educational contexts.

# 1. Introduction

## 1.1 Motivation

Research consistently shows that teaching is one of the most taxing professions in the United States (Agyapong et al., 2022; Creagh et al., 2025). With the highest reported burnout rate of any industry (Doan et al., 2024; Doan et al., 2025; Gallup, 2022), it is no wonder that teacher attrition has become a pressing issue in American education (Carver-Thomas & Darling-Hammond, 2019; Sutcher et al., 2019). Therefore, it is reasonable for educators and educational leaders to turn to emerging AI technology to reduce the intensity of teachers' workload wherever possible. One potential way AI could support teachers is by helping with their planning time, for example, by identifying high-quality instructional materials. Previous research has shown that high-cognitive-demand mathematics tasks have a significant, positive impact on student learning (Stein & Lane, 1996), but most students experience such tasks too rarely (Henningsen & Stein, 1997; Stein et al., 1996), in part because teachers struggle to identify and maintain the cognitive demand of such tasks for their classroom contexts (Boston & Smith, 2009; Stein et al., 1996). However, little is known about AI's capability to reliably identify high-quality mathematics instructional materials. Correctly identifying materials is a challenge for many teachers, and it is therefore possible that AI will also struggle with this task.

In the present study, we built upon previous work (Fox et al., 2026) investigating AI tools' ability to classify math tasks of varying cognitive demand. This prior effort demonstrated that both general AI and AI optimized for mathematics education struggled to correctly identify the cognitive demand level of given tasks. That study used simple prompts, supported only by a published expert guide on the definitions of cognitive demand levels that teachers would likely use to approximate the typical results they would obtain. Here, we examine whether model improvements or prompt optimization will produce better results.

We narrowed our focus to two widely accessible AI tools: one general-purpose (Google Gemini) and one domain-specific (Coteach). We chose Gemini due to its performance on benchmark math evaluations as measured by third-party labs (Arena, 2026; Wang et al., 2024) and Coteach for its relatively strong demonstrated capability in our previous study (Fox et al., 2026). We examined these tools' performance in classifying math tasks of varying cognitive demand across model upgrades and using few-shot prompting methods with an established framework.

## 1.2 Evolving AI Capabilities

Artificial intelligence tools, particularly large language models (LLMs), are evolving rapidly (Rahman, 2024). While model providers regularly release updated versions with general performance improvements or newer capabilities, performance across various domains does not always improve or remain stable, especially within the math domain (Chen et al., 2023). Such local declines can occur due to optimization trade-offs: developers make a model generally more effective or more efficient, but the model loses some of its deeper intelligence capabilities in the process (Dong et al., 2026). Furthermore, within-model performance can decline over time due to a number of factors including changes in the underlying training data, a phenomenon known as temporal drift. For example, OpenAI's GPT-4 had an 84% success rate at identifying prime vs. composite numbers in March 2023, but in June of 2023, GPT-4 identified the same set of numbers with only 51% accuracy. In contrast, the older version, GPT-3.5, performed better on the same task in June 2023 than in March of the same year (Chen et al., 2023). The severity of model degradation depends on the domain and the time between prompts; even short intervals (e.g., 3 months) can result in dramatically different outcomes (Gupta et al., 2025; Tschisgale & Wulff, 2026).

In educational contexts, this temporal instability poses practical problems. A teacher or administrator who selects an AI tool based on research findings or published benchmark performance may find that its capabilities for some educational tasks have improved, while its performance on others has declined. Understanding the temporal stability of AI performance on educational tasks is, therefore, critical for making informed decisions about AI integration into teachers' planning routines and classroom use.

Compounding this instability, the field of prompt engineering has emerged, demonstrating that prompt structure and wording can significantly influence model outputs (Schulhoff et al., 2024). A prompt (i.e., input or request to an AI tool used to inform the output) can vary in terms of the type of input, including text (e.g., What is the difference between a cyclone and a tornado?), code (i.e., programming languages), documents (e.g., PDFs), images (e.g., screenshots), as well as audio and video recordings, and structured datasets (e.g., spreadsheets). Prompts can also vary by the type of task being requested (e.g., problem solving, analysis, evaluation) and use more than one input type to support the AI tool's execution of the task. Critically, while temporal drift operates at the level of the model, reflecting changes in underlying capabilities, prompt drift describes what that instability looks like from the user's perspective: the same prompt, submitted to a newer or updated model, may produce meaningfully different outputs even when nothing about the user's input has changed. In other words, temporal drift is the cause; prompt drift is the downstream consequence that users experience. This distinction matters for educators, who may not realize that a previously reliable prompt has become less effective not because of anything they did, but because the underlying model has changed. Depending on the requested task and the prompting format, prompting techniques may need to change to accommodate a newer AI model's strengths, as some models are better at handling specific input types (e.g., programming language) and requests (e.g., evaluating assignments), and these strengths fluctuate regularly (Chen et al., 2023; Gupta et al., 2025; Schulhoff et al., 2024).

AI models' performance across different input types and requests is regularly tested across domains using benchmark tasks, with results published on websites such as Arena.ai and HuggingFace.co. Subsequently, many prompt engineering techniques (i.e., frameworks for improving prompts) have emerged and have been shown to greatly improve model performance (Schulhoff et al., 2024). Few-shot prompting (sometimes referred to as few-shot learning) is the practice of providing an AI model with a prompt and examples to support in-context learning (Brown et al., 2020). This technique has consistently yielded better results than simple prompting without examples (i.e., zero-shot; Brown et al., 2020; Chen et al., 2023; Schulhoff et al., 2024) and has been used to improve consistency of model outputs (Schulhoff et al., 2024).

In addition to producing differing results over time, results from prompting techniques vary across generalized and specialized tools. For instance, few-shot prompting has been shown to produce better results when used with specialized AI tools than with larger, general-purpose models such as ChatGPT (Chen et al., 2024). This finding could be especially important for teachers who will increasingly work with domain- or curriculum-specific AI tools. Unlike zero-shot prompting (i.e., providing no examples) or extensive fine-tuning (i.e., providing hundreds to thousands of examples to tune model parameters), few-shot prompting allows teachers to use the curricular materials they have on hand as exemplars. Thus, few-shot prompting represents an accessible middle ground for teachers to improve their AI model's output.

While some research has investigated how AI tools perform within the math domain over time, little is known about the relative magnitude of improvement from prompt engineering versus model updates, particularly for evaluating specialized educational math tasks. Specifically, using an AI tool as an evaluator (see Schulhoff et al., 2024) and employing few-shot prompting has not been systematically evaluated for analyzing educational math tasks. It is unknown whether providing AI tools with pre-classified math task exemplars improves classification accuracy, how many exemplars are optimal, or whether improvements persist across different task types. The present study addresses this gap by examining whether few-shot prompting improves AI tools' ability to make fine-grained distinctions in math task classification using the Task Analysis Guide (see Appendix A).

## 1.3 Research Gaps and Research Questions

Most of the current research around AI in education has focused on simulated scenarios with pre-service teachers (Aqazade et al., 2025; Kim et al., 2026; Lee & Yeo, 2022; Lee et al., 2025; Pelton & Pelton, 2024; Son et al., 2024; Uygun et al., 2025; Wijaya et al., 2025) or student-facing uses like intelligent tutoring systems (Hwang & Tu, 2021), leaving critical questions unanswered. The present study addresses the following research questions associated with gaps in research:

**Temporal stability**:

- How stable is AI performance in classifying educational tasks over time? Do model updates improve, maintain, or degrade performance in identifying the cognitive demand of math tasks?
- Do general-purpose and education-specific AI tools show different patterns of temporal stability?

**Few-Shot Prompting**:

- Does few-shot prompting with exemplar tasks improve classification accuracy?
- Does few-shot prompting impact general-purpose and education-specific AI tools differently?
- Does few-shot prompting improve performance uniformly across all cognitive demand categories, or does it perform differently with certain categories?

## 1.4 Theoretical Framework

This study employs the same Task Analysis Guide (TAG) used in our prior research (Fox et al., 2026). Developed by Stein and Smith (1998), the TAG is an evidence-based framework used to support teacher professional development and categorizes mathematical tasks into four levels of cognitive demand:

**Low Cognitive Demand:**

1. **Memorization**: Tasks that require reproduction of previously learned facts, rules, formulas, or definitions without connections to concepts or meaning (e.g., *What are the decimal and percent equivalents for the fractions ½ & ¼?*)
2. **Procedures without Connections**: Tasks that focus on producing correct answers using procedures without developing conceptual understanding (e.g., *Convert the fraction 3/8 to a decimal and a percent. Show your work.*)

**High Cognitive Demand:**

3. **Procedures with Connections**: Tasks that focus students' attention on the use of procedures for the purpose of developing a deeper understanding of mathematical concepts and ideas (e.g., *Using a 10 x 10 grid, identify the decimal and percent equivalents of 3/8.*)
4. **Doing Mathematics**: Tasks that require complex, non-algorithmic thinking and demand self-monitoring of one's own cognitive processes (e.g., *Shade 6 of the small squares in the rectangle below.*

*Using the diagram, explain how to determine each of the following:*

- *the percent of the area that is shaded*
- *the decimal part of the area that is shaded*
- *the fractional part of the area that is shaded)*

We continue to use this framework to enable direct comparison with baseline performance and to examine how time, model updates, and prompting techniques influence performance on this assessment task.

## 1.5 Model Selection Strategy

Rather than testing all the tested tools from our baseline study, we selected two models representing different selection criteria:

**Gemini**: Selected based on benchmark performance 7 weeks after baseline testing. A third-party evaluation using the Massive Multitask Language Understanding Pro benchmark (Wang et al., 2024) demonstrated that Gemini achieved 98% accuracy on mathematics- and psychology-related tasks. This suggested strong domain-relevant capabilities. In our baseline study, Gemini demonstrated an average performance accuracy of 58%.

**Coteach**: Selected based on empirical performance in our baseline study (Fox et al., 2026) where Coteach demonstrated the highest classification accuracy (75%) among education-specific AI tools, suggesting practical effectiveness for our task despite benchmark performance data being unavailable.

This selection strategy allows us to compare tools chosen by different criteria, theoretical capability on domain-relevant tasks (benchmarks) versus demonstrated performance in our previous study (empirical testing), and to examine whether general-purpose versus education-specific tools show different response patterns after model updates and with different prompting techniques.

# 2. Methods and Results

## 2.1 Study Design Overview

This study contrasts relative performance in two models across time points and prompts:

- **Time 1—Baseline (Zero-shot):** Initial testing with standard prompting (task to be evaluated, brief instructions, attached coding framework)
- **Time 2a—7 weeks after Time 1 (Zero-shot)**: Retesting after model updates with standard prompting. Gemini went through major updates (Gemini 2.5 Flash → Gemini 3 Flash). Coteach version updates were not published during this time period but Coteach is regularly updated based upon changes to the underlying model (Anthropic's Claude) and education-specific training materials.

- **Time 2b—7 Weeks after Time 1 (Few-Shot Prompting):** Testing with the addition of few-shot prompting using 2 coded exemplar tasks per category.

## 2.2 Tasks and Materials

The present study used the same twelve mathematics tasks (A–L) employed in our baseline study (see Appendix B for complete task information). All tasks were pre-classified by human experts trained in using the Task Analysis Guide.

**Memorization:** Tasks A, E, L

**Example: Task A**

*What are the decimal and percent equivalents for the fractions ½ and ¼?*

**Procedures Without Connections:** Tasks D, H, J

**Example: Task D**

*Convert the fraction 3/8 to a decimal and a percent. Show your work.*

**Procedures With Connections:** Tasks B, G, I

**Example: Task G**

*Alazar Electric company sells light bulbs to big box stores – the big chain stores that frequently buy large numbers of bulbs in one sale. They sample their bulbs for defects.*

*A sample of 96 light bulbs consisted of 4 defective ones. Assume today's batch of 6,000 light bulbs has the same proportion of defective bulbs as the sample.*

*Set up a proportion and solve it to determine the total number of defective bulbs.*

*The big businesses they sell to accept no larger than a 4% rate of defective bulbs.*

*Is today's batch less than 4% defective? Show your work.*

**Doing Mathematics:** Tasks C, F, K

**Example: Task K**

*This past summer, you were hired to work at Custom T-Shirts. When a customer places an order for a special design, Custom T-Shirts charges a one-time fee of $15 to set up a t-shirt design, plus $8 for each t-shirt printed.*

*What equation can be used to determine how much to charge a customer for any number of shirts? Explain how you determined your answer.*

## 2.3 Prompting Conditions

### *2.3.1 Zero-Shot Prompting (Times 1 and 2a)*

Models received the Task Analysis Guide and the task to be classified with this prompt template:

> *"Based on the 'TAG.docx' I uploaded, determine whether the math task 'Task[X].docx' uploaded is high-cognitive demand or low-cognitive demand. Give detailed reasoning and output the detailed analysis with dimensions in a table."*

Each task in the test set was uploaded one at a time using this prompt to reduce processing strain on the AI tool. This maintained our "out of the box" baseline prompting approach from our previous study (Fox et al., 2026).

### *2.3.2 Few-Shot Prompting (Time 2b)*

For few-shot prompting, we selected two exemplar tasks for each of the four cognitive demand categories (eight exemplars in total). Prior work in the field of prompt engineering has shown that the number of exemplars and their labeling can improve model performance, but only up to about 20 exemplars, after which performance tends to plateau (Schulhoff et al., 2024). The present study used two exemplars per cognitive-demand category to establish a baseline for the utility of few-shot prompting that would not be especially burdensome for educators. Exemplars were selected from a separate pool of middle school math tasks that had been categorized by human experts prior to use in this study (for complete information on the exemplar tasks, see Appendix C). Previous research in prompt engineering has shown that using exemplars conceptually relevant to the test set helps orient AI tools to nuances in classification criteria rather than simply pattern-matching on surface-level wording (Liu et al., 2021; Schulhoff et al., 2024). Thus, the exemplars used in the present study were chosen for their semantic similarity to the tasks in our test set (i.e., they addressed comparable math content and levels of cognitive demand but were not identical in wording or structure).

### *2.3.3 Prompt Engineering Techniques:*

1. **Format conversion**: We formatted the Task Analysis Guide and all exemplar tasks using LaTeX to reduce processing demands associated with long text inputs, as input structure has been shown to affect model performance (Schulhoff et al., 2024).
2. **Exemplar presentation:** Each exemplar was presented with the complete task and the correct classification of the task. The task order presentation followed a low-to-high cognitive demand alignment with the TAG (*Memorization* to *Doing Mathematics*; i.e., not randomized)[1].
3. **Few-shot prompt template:**
   *"Based on the task analysis guide and examples I provided in 'TAG_Task_ex.tex', determine whether the math task 'Task_x.docx' uploaded is high-cognitive demand or low-cognitive demand. Give detailed reasoning and output the detailed analysis with dimensions in a table."*

The combined file (TAG_Task_ex.tex) included the complete Task Analysis Guide followed by the eight exemplar tasks with their classifications. Each task in the test set was uploaded individually using the same prompt to reduce processing strain on the AI tool.

## 2.4 Temporal Stability: Impact of Time and Model Updates

### *2.4.1 Methods*

---

[1] Exemplar presentation order has been shown to meaningfully affect AI classification accuracy; randomized or optimized ordering has been associated with performance gains (Schulhoff et al., 2024).

At Time 1 (baseline, October 30, 2025), both models classified all twelve tasks using standard zero-shot prompting. At Time 2a, seven weeks later, we verified that Google Gemini had undergone a version update (from 2.5 Flash to 3 Flash) and re-administered the same prompting procedure for all twelve tasks. This allowed within-tool comparison of classification accuracy before and after the elapsed time period and, for Gemini, a known model update.

### *2.4.2 Results*

The two models showed notably different temporal stability patterns over the seven-week interval, despite both being subject to elapsed time and potential model changes (see Table 1).

Gemini's overall accuracy was the same at both time points (58%, 7 of 12 tasks). However, even though the overall accuracy was the same, Gemini’s classification of individual tasks shifted. Three tasks that Gemini had correctly classified at Time 1 were misclassified at Time 2a, while three previously misclassified tasks became correct. This pattern of consistent overall accuracy with redistribution of correct responses is consistent with prompt drift (i.e., the model is not behaving the same way, even though its average performance suggests it is). This distinction matters for practitioners who rely on published benchmark accuracy ratings to choose an AI tool; a tool that performs at the same overall level on a benchmark may behave differently on the specific tasks they care about.

Coteach showed a different pattern: a 25-percentage-point decline in overall accuracy (75% to 50%), due to losing accuracy on four tasks it had previously classified correctly. Unlike Gemini, this represents a substantial loss of capability rather than a redistribution of correct responses. Because Coteach's model version information is not publicly available, we cannot determine whether this decline reflects an underlying model update, prompt drift, or some combination of both.

Together, these patterns illustrate two distinct ways that AI tools can become less reliable for educators over time: visible degradation (Coteach), where overall accuracy clearly drops, and invisible drift (Gemini), where overall accuracy is preserved but task-level behavior shifts in ways that benchmark reports would not detect. For complete task-by-task results, see Table 1 (Section 2.7).

## 2.5 Few-Shot Prompting

### *2.5.1 Methods*

Immediately following the Time 2a zero-shot assessment, we administered the few-shot prompting condition to both models for all twelve tasks. Tasks were uploaded one at a time, and the prompt was repeated to reduce processing demands. This allowed direct comparison of prompting approaches within the same model versions.

### *2.5.2 Results*

Few-shot prompting improved classification accuracy for both models (see Table 1). For Gemini, overall accuracy increased from 58% to 67% (8 of 12 tasks). For Coteach, accuracy increased from 50% to 75% (9 of 12 tasks), representing a full return to baseline performance. Notably, the few-shot improvement for Coteach exactly offset the decline observed from Time 1 to Time 2a, suggesting that few-shot prompting can compensate for performance losses attributable to temporal drift.

## 2.6 Analysis

Comparing the magnitude of change across conditions revealed that few-shot prompting was more effective than model updates. For Gemini, few-shot prompting produced larger gains (+8 percentage points) than the model update alone (+/- 0). For Coteach, the model showed degraded performance between Time 1 and Time 2a but recovered after few-shot prompting. Together, these findings suggest that prompt optimization is a viable solution for combatting temporal drift.

## 2.7 Category-Specific Effects

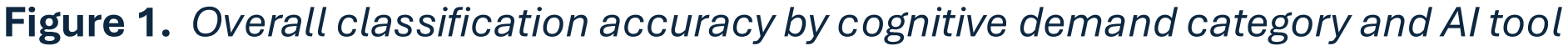

**Figure 1.** *Overall classification accuracy by cognitive demand category and AI tool*

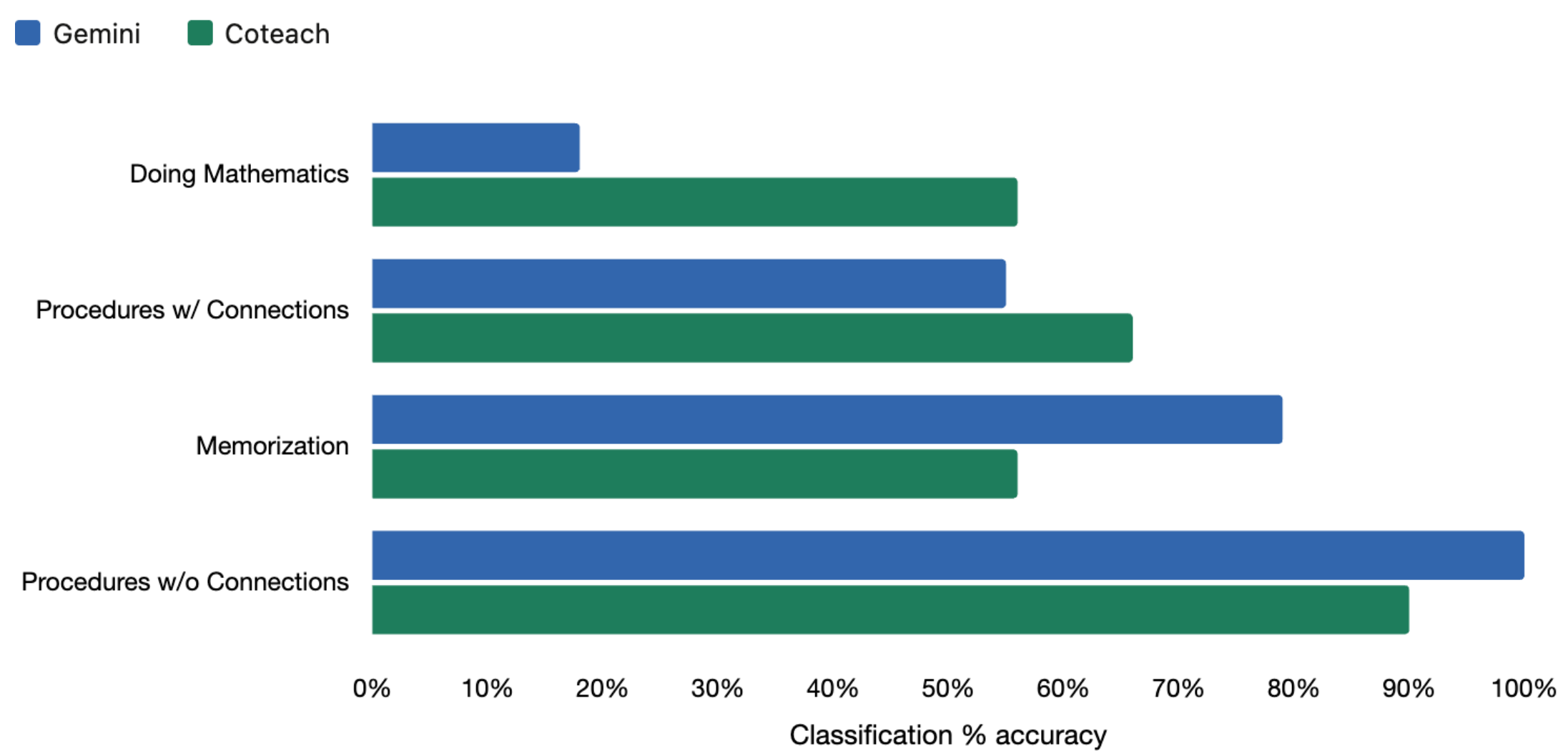

Accuracy varied substantially by cognitive demand category. *Procedures without Connections* was the category classified most accurately (avg. 94%), while *Doing Mathematics* was the least (avg. 33%). *Memorization* and *Procedures with Connections* fell in between at 67% and 61%, respectively. These findings diverge slightly from our previous work (Fox et al., 2026), which found higher accuracy for middle categories than the extremes.

Across tasks, several recurring patterns emerged (see Table 1): some tasks were classified consistently correctly across all conditions; others showed a U-shaped pattern (correct at T1, incorrect at T2a, correct again at T2b); one task (G) was resistant to correction by either model update or few-shot prompting; and *Doing Mathematics* tasks showed the most variability, with some tasks remaining misclassified regardless of condition.

**Table 1.** *Task classification accuracy across task, time, and AI tool*

Correct · Incorrect | Time 1 = baseline · Time 2a = zero-shot · Time 2b = few-shot

| | Time 1 | | Time 2a | | Time 2b | |
|---|---|---|---|---|---|---|
| Task | Gemini | Coteach | Gemini | Coteach | Gemini | Coteach |
| Memorization (avg. 67%) | | | | | | |
| A | ✓ | ✓ | ✗ | ✗ | ✓ | ✓ |
| E | ✓ | ✗ | ✓ | ✗ | ✓ | ✗ |
| L | ✗ | ✓ | ✓ | ✓ | ✓ | ✓ |
| Procedures without Connections (avg. 94%) | | | | | | |
| D | ✓ | ✓ | ✓ | ✓ | ✓ | ✓ |
| H | ✓ | ✓ | ✓ | ✗ | ✓ | ✓ |
| J | ✓ | ✓ | ✓ | ✓ | ✓ | ✓ |
| Procedures with Connections (avg. 61%) | | | | | | |
| B | ✗ | ✓ | ✓ | ✓ | ✓ | ✓ |
| G | ✓ | ✓ | ✗ | ✗ | ✗ | ✗ |
| I | ✓ | ✓ | ✗ | ✗ | ✓ | ✓ |
| Doing Mathematics (avg. 33%) | | | | | | |
| C | ✗ | ✗ | ✗ | ✗ | ✗ | ✗ |
| F | ✗ | ✓ | ✓ | ✓ | ✗ | ✓ |
| K | ✗ | ✗ | ✗ | ✓ | ✗ | ✓ |

## 2.8 Comparing General-Purpose vs. Education-Specific Tools

**Gemini (General-Purpose):**

- Temporal stability (T1 → T2a): overall accuracy remained the same, but different correct/incorrect classifications
- Moderate few-shot improvement (+8% points)
- Achieved a higher final performance (67%) but from a lower starting point

**Coteach (Education-Specific):**

- Temporal instability (notable degradation after 7 weeks)
- Large few-shot improvement (+25% points)
- Returned to strong performance (75%), matching its baseline

These patterns are consistent with previous work (Chen et al., 2024) and with the hypothesis that education-specific tools may be more vulnerable to temporal drift but more responsive to domain-specific prompting techniques, whereas general-purpose tools may be more stable across model updates, but less sensitive to specialized prompting. However, the present study is limited in that we used only one AI tool representative per category (Gemini and Coteach), and the differences in results may be due to unknown factors specific to these tools rather than to systematic differences between tool categories. Future work should investigate this hypothesis across a larger range of general and education-specific AI tools.

## 2.9 Final Performance Rankings

Notably, Coteach's performance surpassed Gemini's performance, even with the degradation between Time 1 and 2a, suggesting that with strategic prompting, education-specific tools can leverage their specialized training.

# 3. Discussion

## 3.1 Interpretation of Findings

This study examined two aspects of AI tool use within educational contexts: temporal stability and the effect of few-shot prompting.

**RQ1 & RQ2: Temporal Stability?** AI performance was not stable over time, even over a relatively short seven-week interval, though the two tools differed in both the nature and severity of their instability. Gemini's overall accuracy remained constant at 58% across the model update from version 2.5 Flash to version 3 Flash, but the tasks it classified correctly shifted, which is a pattern consistent with prompt drift (Chen et al., 2023). This phenomenon is particularly concerning for practitioners because it would not be detected by published benchmark reports and can produce meaningful differences in certain applications. In contrast, Coteach showed a 25-percentage-point drop in overall accuracy, representing degradation rather than redistribution. The internal mechanisms behind this drop cannot be pinpointed because Coteach's model update information is not publicly available; however, the extent and pattern are consistent with a model update, temporal drift, or some combination of both. Taken together, the two tools illustrate two distinct failure modes, task-level instability and degradation. Additionally, Coteach's lack of transparency is an important finding because, without model version information, users are limited in their ability to draw conclusions regarding reliability.

**RQ3 & RQ4: Does few-shot prompting with exemplar tasks improve classification accuracy, and does it affect models differently?** Yes. Few-shot prompting with two exemplars per cognitive-demand category improved Gemini's accuracy by 8 percentage points (58% to 67%) and Coteach's by 25 percentage points (50% to 75%). These results showed that the customized AI tool benefited more from few-shot prompting efforts than the general AI tool, which is a pattern consistent with prior findings that smaller, fine-tuned AI models can outperform larger general-purpose models with prompt engineering (Chen et al., 2024). These results warrant further investigation with a larger task sample to determine if they generalize.

**RQ5: Does few-shot prompting improve performance uniformly across cognitive demand categories?** No. The benefit of few-shot prompting was not evenly distributed. Tasks in the *Procedures without Connections* category had the highest baseline classification accuracy, leaving little room for improvement. This represents a simple pragmatic point: no need to fix what isn't broken. More importantly, tasks in the *Doing Mathematics* category remained low even with few-shot prompting. This is a foundational challenge in how LLMs approach the classification of such tasks that few-shot prompting cannot address.

Our previous work (Fox et al., 2026) found similar results with zero-shot prompting: *Doing Mathematics* tasks were accurately classified only 27% of the time and were almost always misclassified as *Procedures with Connections*. This pattern is consistent with two failure modes: central-tendency bias and shortcut learning. Central-tendency bias occurs when judges avoid extreme classifications or ratings in favor of categories or scores in the middle of a scale, resulting in an underrepresentation of the highest and lowest values. Our prior analysis also highlighted AI tools' failure to distinguish between characteristics that define each cognitive demand category, instead focusing on surface level text features. For example, Task K (*Doing Mathematics*) was misclassified 91% of the time, with AI tools citing the presence of the problem's transparent cost structure as providing a predictable solution path rather than noting the absence of a common algorithmic pathway; the absence is a key feature of *Doing Mathematics* tasks.

AI tools' reliance on surface-level text features (e.g., keywords, lexical matching) that correspond to labels in the training data rather than to deeper semantic structure is a common phenomenon in machine learning known as shortcut learning (Geirhos et al., 2020). AI tools perform well on benchmark tasks but are unable to demonstrate the same level of proficiency in real-world contexts because they amplify spurious cues from prompts and exemplars rather than actually learning a rule (Tang et al., 2023). Furthermore, larger models are more likely to use shortcuts than smaller models (Tang et al., 2023). This helps explain why few-shot prompting did not close the gap for *Doing Mathematics* tasks in the present study, and perhaps why Coteach outperformed Gemini.

Taken together, these findings illustrate that few-shot prompting can offset losses in model capability over time, but not uniformly across task types. Correcting this may require more targeted prompt engineering, for example, chain-of-thought prompting that explicitly directs tools to evaluate what a task does *not* require, or rubric language refined to make the absence-based criteria of *Doing Mathematics* more salient than its procedural-looking surface features.

## 3.2 Future Directions

The present study built upon our previous baseline study, which investigated how accurately AI tools classified the cognitive demand of mathematics tasks when provided with the Task Analysis Guide (Fox et al., 2026). The results have highlighted several areas for future work to address limitations and advance understanding of how AI tools can support educators.

**Limited AI tool sample:** We tested two AI tools, which represent a small fraction of the general and education-specific AI tools currently available. The patterns we encountered may not generalize to other tools. Investigating more models would reveal whether these specific accuracy rates and error patterns are universal or tool-specific. However, the kinds of failures we documented, central-tendency bias, reliance on surface-level features, and the resulting difficulty with *Doing Mathematics* tasks, are not idiosyncratic to any single model. They represent well-documented properties of underlying architecture and training standards that current general-purpose LLMs share (Du et al., 2023; Geirhos et al., 2020; Tang et al., 2023). Education-specific tools, in turn, are typically built on top of these same foundation models (e.g., Coteach on Claude). It is therefore reasonable to expect that the broad patterns reported here would replicate across other current-generation tools, even if the specific accuracy rates vary. Future work testing a wider range of models would help establish the boundaries of this generalization and identify any tools that meaningfully diverge from the pattern.

**Limited tasks and domain:** The twelve tasks we used for cognitive demand classification are regularly used in teacher training and professional development and were representative of each level of cognitive demand outlined by the TAG. And while these tasks could be used in classroom instruction, they do not reflect the wide array of tasks and task formats. Math tasks vary in how they are developed (textbook-embedded, teacher-created, or emerging organically from student questioning), format (open-ended, goal-oriented, inquiry-based), purpose (exploration, instruction, assessment), where they fit within the learning trajectory, the amount of time they take, and the materials they require to complete (Markle & McGarvey, 2025). These differences are important to consider, as classification accuracy may differ systematically across these characteristics. Additionally, the math content in the test set tasks covered only middle-school-level material, raising questions about whether AI tools would perform differently on material from other grade levels. Future work should examine whether gains from few-shot prompting transfers to a broader range of task types and content.

**Short time interval:** Seven weeks between Time 1 and Time 2 represents a relatively brief interval between testing, and longer-term temporal patterns with these tools and math-specific tasks remain unknown. Following models across multiple version updates would reveal longer-term stability patterns. However, given the failure modes we observed and the literature stating that shortcut learning persists across models, there is good reason to expect that the broad patterns documented here will persist even as specific model versions change (Geirhos et al., 2020; Tang et al., 2023). This is not to say that future models will perform identically, but it does suggest that the kinds of errors we identified are unlikely to disappear with incremental version updates, and that the implications of this study for educational practice are not dependent on a specific snapshot in time.

**Few-shot approach:** We tested a specific few-shot prompting strategy that used two exemplars for each cognitive demand category. We chose two exemplars per category based on prior work in prompt engineering, which found the largest performance gains when moving from zero-shot to one- or two-shot prompting, with diminishing returns beyond two exemplars (Brown et al., 2020). AI performance with few-shot prompting is sensitive to the number of exemplars used (Zhao et al., 2021), and thus we kept the number of exemplars low to avoid introducing noise or reinforcing pattern matching. In the future, studies should test different exemplar selection strategies, different exemplar ordering, and different prompting frameworks to systematically determine which combination of exemplars and techniques consistently produces the most accurate results. For example, the order in which exemplars are presented can influence AI tools' performance accuracy dramatically, with the same exemplars in different orders producing differing accuracies (<50% — 90%; Schulhoff et al., 2024).

**Limited model transparency:** We documented what changed (a practitioner goal) but not why (a scientific goal). Here, we focused on the available models that teachers are especially likely to use, which are generally not described transparently. Understanding the mechanisms behind temporal instability and few-shot prompting would require access to model information for Coteach, as well as more detailed information regarding the internal parameters and attention patterns for both AI tools. If model internals become accessible, understanding why updates affect specialized performance would inform better strategies for AI use after updates. Research using open-source models and examining the effects of different model optimization efforts would be useful for the field.

**No cost analysis:** Few-shot prompting increases API token usage and thus may increase cost for users. We did not evaluate whether the accuracy improvements justify increased expenses. Quantifying the trade-offs between accuracy improvements and increased computational costs would guide practical decision-making.

# 4. Conclusion

This study reveals that AI performance on classifying educational tasks is temporally unstable. For Gemini, overall performance accuracy remained stable after a model update but showed task-specific behavioral instability. Model versions were unavailable for Coteach, so the underlying reasons for performance degradation between Time 1 and Time 2a had to be inferred, likely due either to an unknown model update (i.e., prompt drift) or temporal drift. Few-shot prompting with exemplar tasks produced substantial improvements for both models (+8 and +25 percentage points, respectively), demonstrating that active prompt optimization is more reliable and effective than passive reliance on model improvements.

Coteach's performance degraded by 25 percentage points between Times 1 and 2a, and its recovery with few-shot prompting highlights both the risks of prompt/temporal drift and the power of thoughtful prompt engineering. For educational applications, this suggests that investing in prompt optimization may be more productive than waiting for better model versions.

More broadly, this research challenges the assumption that newer AI models are always better for specialized educational tasks. As AI tools become integrated into educational practice, better evaluation frameworks that account for temporal instability and prioritize prompt optimization are needed. The continued evolution of AI tools requires users to update their evaluation, selection, and use within educational contexts.

# References

Agyapong, B., Obuobi-Donkor, G., Burback, L., & Wei, Y. (2022). Stress, burnout, anxiety and depression among teachers: A Scoping Review. *International Journal of Environmental Research and Public Health*, *19*(17), 10706. https://doi.org/10.3390/ijerph191710706

Aqazade, M., Mauntel, M., & Atabas, S. (2025). Empowering mathematics teacher educators: Exploring Artificial Intelligence-driven mathematical tasks. *School Science and Mathematics*, 126(1), 24–43.

Arena. (2026). Arena leaderboard: Compare & benchmark the best frontier AI models. https://arena.ai/leaderboard

Boston, M. D., & Smith, M. S. (2009). Transforming secondary mathematics teaching: Increasing the cognitive demands of instructional tasks used in teachers' classrooms. *Journal for Research in Mathematics Education, 40*(2), 119–156.

Brown, T., Mann, B., Ryder, N., Subbiah, M., Kaplan, J. D., Dhariwal, P., ... & Amodei, D. (2020). Language models are few-shot learners. *Advances in neural information processing systems*, *33*, 1877-1901.

Carver-Thomas, D., & Darling-Hammond, L. (2019). Teacher turnover: Why it matters and what we can do about it. *Learning Policy Institute*. https://doi.org/10.54300/454.278

Chen, E., Wang, D., Xu, L., Cao, C., Fang, X., & Lin, J. (2024). A systematic review on prompt engineering in large language models for K-12 STEM education. *ArXiv*. https://arxiv.org/abs/2410.11123

Chen, L., Zaharia, M., & Zou, J. (2023). How is ChatGPT's behavior changing over time? *ArXiv*. https://arxiv.org/abs/2307.09009

Creagh, S., Thompson, G., Mockler, N., Stacey, M., & Hogan, A. (2025). Workload, work intensification and time poverty for teachers and school leaders: a systematic research synthesis. *Educational Review*, 77(2), 661–680. https://doi.org/10.1080/00131911.2023.2196607

Doan, S., Steiner, E. D., Woo, A., & Pandey, R. (2024). *State of the American Teacher survey: 2024 technical documentation and survey results* (RR-A1108-11). RAND Corporation. https://www.rand.org/pubs/research_reports/RRA1108-11.html

Doan, S., Steiner, E. D., & Pandey, R. (2025). Teacher well-being, pay, and intentions to leave in 2025: Findings from the State of the American Teacher survey (RR-A1108-16). RAND Corporation. https://www.rand.org/pubs/research_reports/RRA1108-16.html

Dong, H., Acun, B., Chen, B., & Chi, Y. (2026). Scalable LLM reasoning acceleration with low-rank distillation (arXiv:2505.07861). arXiv. https://doi.org/10.48550/arXiv.2505.07861

Du, M., He, F., Zou, N., Tao, D., & Hu, X. (2023). Shortcut learning of large language models in natural language understanding. *Communications of the ACM*, 67(1), 110–120.

Fox, D. S., Robles, B. L., Brovey, E. D., & Schunn, C. D. (2026). Baseline performance of AI tools in classifying cognitive demand of mathematical tasks. *ArXiv*. https://arxiv.org/abs/2603.03512

Gallup. (2022). K-12 workers have highest burnout rate in U.S. Gallup News. https://news.gallup.com/poll/393500/workers-highest-burnout-rate.aspx

Geirhos, R., Jacobsen, J. H., Michaelis, C., Zemel, R., Brendel, W., Bethge, M., & Wichmann, F. A. (2020). Shortcut learning in deep neural networks. *Nature Machine Intelligence*, *2*(11), 665-673. https://doi.org/10.1038/s42256-020-00257-z

Gupta, M., Virostko, J., & Kaufmann, C. (2025). Large language models in radiology: Fluctuating performance and decreasing discordance over time. *European Journal of Radiology, 182*, 111842. https://doi.org/10.1016/j.ejrad.2024.111842

Henningsen, M., & Stein, M. K. (1997). Mathematical tasks and student cognition: Classroom-based factors that support and inhibit high-level mathematical thinking and reasoning. *Journal for Research in Mathematics Education, 28*(5), 524–549.

Hwang, G.-J., & Tu, Y.-F. (2021). Roles and research trends of artificial intelligence in mathematics education: A bibliometric mapping analysis and systematic review. *Mathematics, 9*(6), Article 584. https://doi.org/10.3390/math9060584

Kim, Y. R., Park, M. S., & Joung, E. (2026). Exploring the integration of artificial intelligence in math education: Preservice teachers' experiences and reflections on problem-posing activities with ChatGPT. *School Science and Mathematics*, *126*(1), 9-23.

Lee, D., & Yeo, S. (2022). Developing an AI-based chatbot for practicing responsive teaching in mathematics. *Computers & Education*, 191, 104646.

Lee, D., Son, T., & Yeo, S. (2025). Impacts of interacting with an AI chatbot on preservice teachers' responsive teaching skills in math education. *Journal of Computer Assisted Learning*, 41(1), e13091.

Liu, J., Shen, D., Zhang, Y., Dolan, B., Carin, L., & Chen, W. (2021). What makes good in-context examples for GPT-3? (arXiv:2101.06804). arXiv. https://doi.org/10.48550/arXiv.2101.06804

Markle, J., & McGarvey, L. M. (2025). Research on task design in pre-service mathematics teacher education: A scoping review. *Frontiers in Education*, *10*, 1467482. https://doi.org/10.3389/feduc.2025.1467482

Pelton, T., & Pelton, F. L. (2024). Using generative AI in mathematics education: Critical discussions and practical strategies for preservice teachers, teachers, and teacher educators. In J. Cohen & G. Solano (Eds.), *Proceedings of Society for Information Technology & Teacher Education International Conference* (pp. 1800-1805).

Rahman, R. (2024). *The pace of large-scale model releases is accelerating*. Epoch AI. https://epoch.ai/data-insights/large-scale-model-releases

Schulhoff, S., Ilie, M., Balepur, N., Kahadze, K., Liu, A., Si, C., Li, Y., Gupta, A., Han, H., Schulhoff, S., Dulepet, P. S., Vidyadhara, S., Ki, D., Agrawal, S., Pham, C., Kroiz, G., Li, F., Tao, H., Srivastava, A., . . . Resnik, P. (2024). The prompt report: A systematic survey of prompting techniques (arXiv:2406.06608). *ArXiv*. https://doi.org/10.48550/arXiv.2406.06608

Son, T., Yeo, S., & Lee, D. (2024). Exploring elementary preservice teachers' responsive teaching in mathematics through an artificial intelligence-based Chatbot. *Teaching and Teacher Education*, 146, 104640.

Stein, M. K., Grover, B. W., & Henningsen, M. (1996). Building student capacity for mathematical thinking and reasoning: An analysis of mathematical tasks used in reform classrooms. *American Educational Research Journal, 33*(2), 455–488.

Stein, M. K., & Lane, S. (1996). Instructional tasks and the development of student capacity to think and reason: An analysis of the relationship between teaching and learning in a reform mathematics project. *Educational Research and Evaluation*, *2*(1), 50-80.

Stein, M. K., & Smith, M. S. (1998). Mathematical tasks as a framework for reflection: From research to practice. *Mathematics Teaching in the Middle School*, 3(4), 268-275.

Sutcher, L., Darling-Hammond, L., & Carver-Thomas, D. (2019). Understanding teacher shortages: An analysis of teacher supply and demand in the United States. *Education Policy Analysis Archives, 27*(35).

Tang, R., Kong, D., Huang, L., & Xue, H. (2023). Large language models can be lazy learners: Analyze shortcuts in in-context learning. *Findings of the Association for Computational Linguistics: ACL 2023*, 4645–4657.

Tschisgale, P., & Wulff, P. (2026). Daily and weekly periodicity in large language model performance and its implications for research. *ArXiv*. https://arxiv.org/abs/2602.15889

Uygun, T., Şendur, A., Top, B., & Coşgun-Başeğmez, K. (2025). Facilitating the development of preservice teachers' geometric thinking through artificial intelligence (AI) assisted augmented reality (AR) activities: The case of platonic solids. *Education and Information Technologies*, *30*(7), 8373-8411.

Wang, Y., Ma, X., Zhang, G., Ni, Y., Chandra, A., Guo, S., ... & Chen, W. (2024). Mmlu-pro: A more robust and challenging multi-task language understanding benchmark. *Advances in Neural Information Processing Systems*, *37*, 95266-95290.

Wijaya, T. T., Su, M., Cao, Y., Weinhandl, R., & Houghton, T. (2025). Examining Chinese preservice mathematics teachers' adoption of AI chatbots for learning: Unpacking perspectives through the UTAUT2 model. *Education and Information Technologies*, *30*(2), 1387-1415.

Zhao, Z., Wallace, E., Feng, S., Klein, D., & Singh, S. (2021). Calibrate before use: Improving few-shot performance of language models. *Proceedings of the 38th International Conference on Machine Learning*, 139, 12697–12706.

# Appendices

**Appendix A: Mathematical Task Analysis Guide**

| **Lower-Level Demands** | **Higher-Level Demands** |
|---|---|
| *Memorization Tasks*<br>• Involves either producing previously learned facts, rules, formulae, or definitions OR committing facts, rules, formulae, or definitions to memory.<br>• Cannot be solved using procedures because a procedure does not exist or because the time frame in which the task is being completed is too short to use a procedure.<br>• Are not ambiguous – such tasks involve exact reproduction of previously seen material and what is to be reproduced is clearly and directly stated.<br>• Have no connection to the concepts or meaning that underlie the facts, rules, formulae, or definitions being learned or reproduced. | *Procedures with Connections Tasks*<br>• Focus students' attention on the use of procedures for the purpose of developing deeper levels of understanding of mathematical concepts and ideas.<br>• Suggest pathways to follow (explicitly or implicitly) that are broad general procedures that have close connections to underlying conceptual ideas as opposed to narrow algorithms that are opaque with respect to underlying concepts.<br>• Usually are represented in multiple ways (e.g., visual diagrams, manipulatives, symbols, problem situations). Making connections among multiple representations helps to develop meaning.<br>• Require some degree of cognitive effort. Although general procedures may be followed, they cannot be followed mindlessly. Students need to engage with the conceptual ideas that underlie the procedures in order to successfully complete the task and develop understanding. |
| *Procedures without Connections Tasks*<br>• Are algorithmic. Use of the procedure is either specifically called for, or its use is evident based on prior instruction, experience, or placement of the task.<br>• Require limited cognitive demand for successful completion. There is little ambiguity about what needs to be done and how to do it.<br>• Have no connection to the concepts or meaning that underlie the procedure being used.<br>• Are focused on producing correct answers rather than developing mathematical understanding.<br>• Require no explanations, or explanations that focus solely on describing the procedure that was used. | *Doing Mathematics Tasks*<br>• Requires complex and non-algorithmic thinking (i.e., there is not a predictable, well-rehearsed approach or pathway explicitly suggested by the task, task instructions, or a worked-out example).<br>• Requires students to explore and to understand the nature of mathematical concepts, processes, or relationships.<br>• Demands self-monitoring or self-regulation of one's own cognitive processes.<br>• Requires students to access relevant knowledge and experiences and make appropriate use of them in working through the task.<br>• Requires students to analyze the task and actively examine task constraints that may limit possible solution strategies and solutions.<br>• Requires considerable cognitive effort and may involve some level of anxiety for the student due to the unpredictable nature of the solution process required. |

**Appendix B: Mathematics Tasks A-L**

**Task A:**
What are the decimal and percent equivalents for the fractions ½ and ¼?

**Task B:**
Using a 10 x 10 grid, identify the decimal and percent equivalents of 3/5.

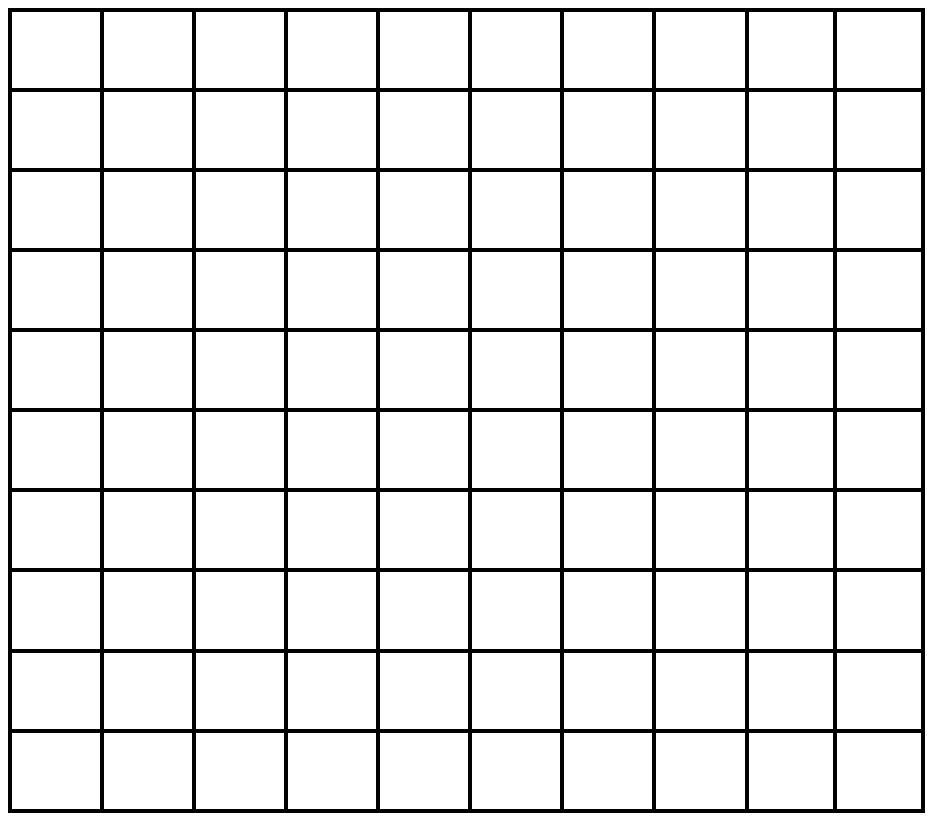

**Task C:**
Shade 6 of the small squares in the rectangle below.

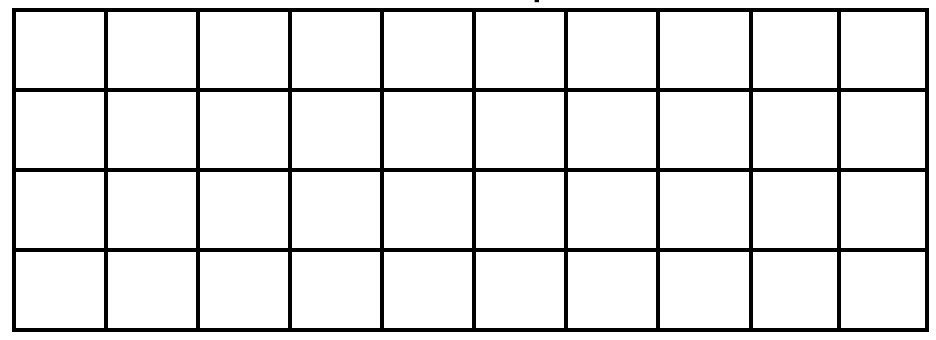

Using the diagram, explain how to determine each of the following:

- the percent of the area that is shaded
- the decimal part of the area that is shaded
- the fractional part of the area that is shaded

**Task D:**
Convert the fraction 3/8 to a decimal and a percent. Show your work.

**Task E:**
True or False? 4/100 < 4/96

**Task F:**
Alazar Electric company sells light bulbs to big box stores – the big chain stores that frequently buy large numbers of bulbs in one sale. They sample their bulbs for defects.
A sample of 96 light bulbs consisted of 4 defective ones. Assume today's batch of 6,000 light bulbs has the same proportion of defective bulbs as the sample.

- What are the total number of defective bulbs made today?

The big businesses agree to accept no larger than a 4% rate of defective bulbs.
Does today's batch meet that expectation? Explain in writing and with equations how you made your decision.

**Task G:**
Alazar Electric company sells light bulbs to big box stores – the big chain stores that frequently buy large numbers of bulbs in one sale. They sample their bulbs for defects.
A sample of 96 light bulbs consisted of 4 defective ones. Assume today's batch of 6,000 light bulbs has the same proportion of defective bulbs as the sample.

- Set up a proportion and solve it to determine the total number of defective bulbs.

The big businesses they sell to accept no larger than a 4% rate of defective bulbs.
Is today's batch less than 4% defective? Show your work.

**Task H:**
Use cross products to solve the proportion. Show your work.
$4/96 = x/6000$

**Task I:**
The table below shows the price in dollars that Custom T-Shirts charges a customer for a given number of t-shirts.
What equation can be used to determine how much to charge a customer for any number of shirts? Explain how you determined your answer.

| Number of Shirts | 1 | 2 | 3 | 4 | 5 |
|---|---|---|---|---|---|
| Price in Dollars | 23 | 31 | 39 | 47 | 55 |

**Task J:**
If $y = 8x + 15$, evaluate y when x =

A. 10
B. 20
C. 30

Explain the procedure you used to find your solution.

**Task K:**
This past summer, you were hired to work at Custom T-Shirts. When a customer places an order for a special design, Custom T-Shirts charges a one-time fee of $15 to set up a t-shirt design, plus $8 for each t-shirt printed.

What equation can be used to determine how much to charge a customer for any number of shirts? Explain how you determined your answer.

**Task L:**
In the following equations, name the slope and y-intercept.

A. $y = 8x + 15$
B. $y = -2x + 7$
C. $y = 4x - 9$

## Appendix C: Exemplar Tasks

**Task 1: Memorization**

Complete the following multiplication facts in one minute or less.

| | |
|---|---|
| 22 × 3 = | 7 × 9 = |
| 4 × 7 = | 8 × 7 = |
| 9 × 5 = | 10 × 6 = |
| 6 × 8 = | 8 × 4 = |
| 3 × 9 = | 5 × 5 = |
| 5 × 4 = | 2 × 6 = |
| 8 × 10 = | 9 × 2 = |
| 3 × 4 = | |

**Task 2: Memorization**

Match the property name with the appropriate equation.

| Properties | Equations |
|---|---|
| 1. Commutative property of addition | a. r(s + t) = rs + rt |
| 2. Commutative property of multiplication | b. x· 1/x = 1 |
| 3. Associative property of addition | c. –y + x = x + (–y) |
| 4. Associative property of multiplication | d. a/b + –a/b = 0 |
| 5. Identity property of addition | e. y· (zx) = (yz)· x |
| 6. Identity property of multiplication | f. 1· (xy) = xy |
| 7. Inverse property of addition | g. d· 0 = 0 and 0· d = 0 |
| 8. Inverse property of multiplication | h. x + (b + c) = (x + b) + c |
| 9. Distributive property | i. y + 0 = y |
| 10. Property of zero for multiplication | j. p· q = q· p |

**Task 3: Procedures without Connections**

Manipulatives or Tools Available: One triangle pattern block

Task: Using the edge of a triangle pattern block as the unit of measure, determine the perimeter of the following pattern-block trains.

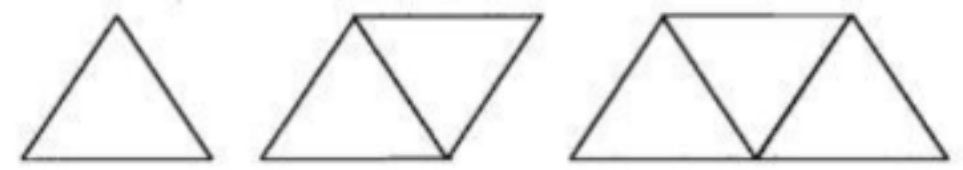

**Task 4: Procedures without Connections**

Write and solve a proportion for each of the following:

| | |
|---|---|
| 17 is what percent of 68? | What is 60% of 45? |
| What is 15% of 60? | 36 is what percent of 90? |
| 8 is 10% of what number? | What is 80% of 120? |
| 24 is 25% of what number? | 21 is 30% of what number? |
| 28 is what percent of 140? | |

**Task 5: Procedures with Connections**

Anita has four 20-point projects for science class. Her scores are shown below. What is her average score? Find the average for Anita's scores by leveling off the stacks.

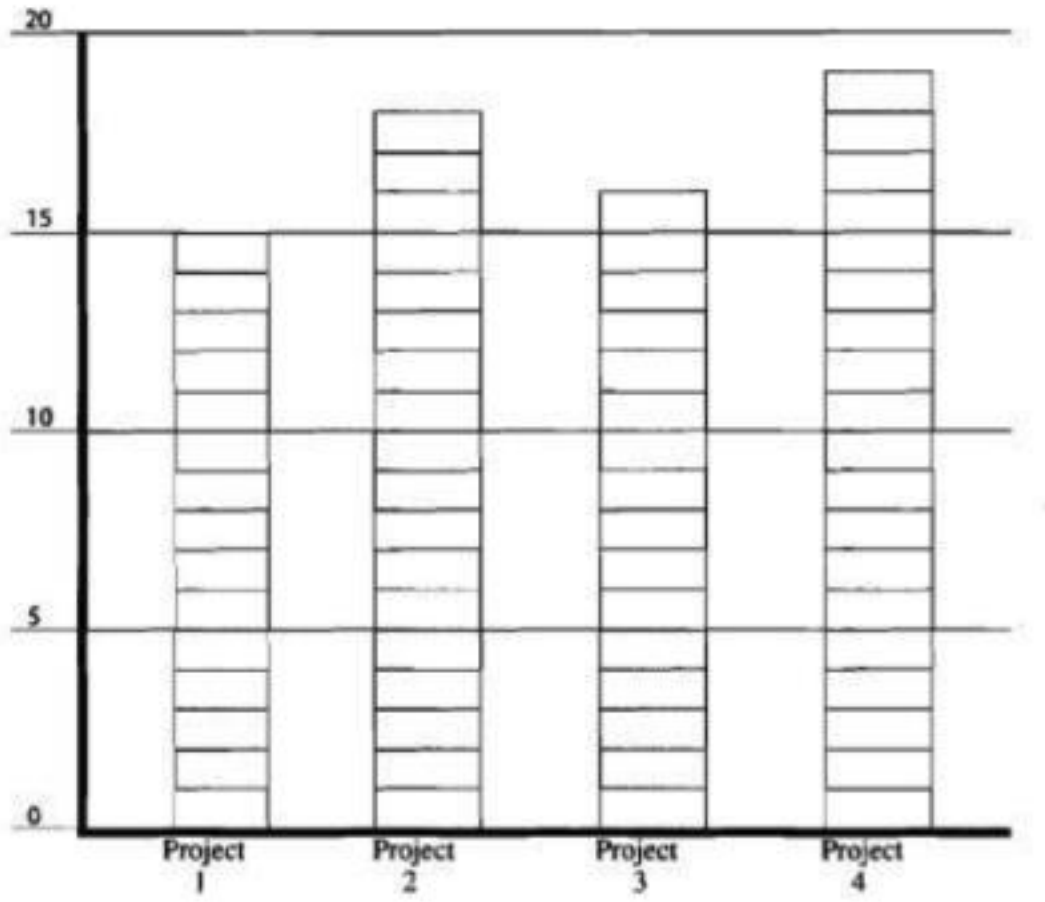

**Task 6: Procedures with Connections**

Part A: The place kicker on the North High School football team has made 13 out of 20 field goals so far this season. The place kicker on the South High football team has made 15 out of 25 field goals so far this season. Which player has made the greatest percentage of field goals?

Part B: If the "better" player does not play for the rest of the season, how many field goals would the other player have to make in the next 10 attempts to have the greatest percentage of field goals?

**Task 7: Doing Mathematics**

Manipulatives or Tools Available: Grid paper, interlocking cubes

1. The kindergarten class is coming to watch a play in our classroom. There are 20 students. In what different ways could we arrange the chairs for them so that all the rows are equal?
2. The two third-grade classes are going to watch our play in the cafeteria. There are 49 students altogether. In what different ways would we arrange the chairs for them so that all the rows are equal?
3. What do you notice about your solutions for problem 1 and problem 2?

**Task 8: Doing Mathematics**

Manipulatives or Tools Available: Calculator

Treena won a 7-day scholarship worth $1,000 to the Pro Shot Basketball Camp. Round-trip travel expenses to the camp are $335 by air or $125 by train. At the camp she must choose between a week of individual instruction at $60 a day or a week of group instruction at $40 a day. Treena's food and other expenses are fixed at $45 a day. If she does not plan to spend any money other than the scholarship, what are all choices of travel and instruction plans she could afford to make? Explain which option you think Treena should select and why.